\documentclass[runningheads]{llncs}
\usepackage{graphicx}
\usepackage{amsmath}
\usepackage{amssymb}
\usepackage{multirow}
\usepackage{float}
\usepackage{algorithm}
\usepackage{algpseudocode}
\usepackage{hyperref}
\usepackage{algpseudocode}
\usepackage{tabularx, array,booktabs}
\usepackage{tikz}
\usetikzlibrary{spy}
\usepackage{comment}
\usepackage{xcolor}

\begin{document}

\title{DHFormer: A Vision Transformer-Based Attention Module for Image Dehazing}
%
%
\author{Abdul Wasi\inst{1} \and
O.Jeba Shiney\inst{1}}
\authorrunning{Abdul Wasi et al.}
%
\institute{Chandigarh University, Mohali, India
\\
\email{wasilone11@gmail.com}
\\
\email{jeba.e10900@cumail.in}}
\maketitle              
\begin{abstract}
Images acquired in hazy conditions have degradations induced in them. Dehazing such images is a vexed and ill-posed problem. Scores of prior-based and learning-based approaches have been proposed to mitigate the effect of haze and generate haze-free images. Many conventional methods are constrained by their lack of awareness regarding scene depth and their incapacity to capture long-range dependencies. In this paper, a method that uses residual learning and vision transformers in an attention module is proposed. It essentially comprises two networks: In the first one, the network takes the ratio of a hazy image and the approximated transmission matrix to estimate a residual map. The second network takes this residual image as input and passes it through convolution layers before superposing it on the generated feature maps. It is then passed through global context and depth-aware transformer encoders to obtain channel attention. The attention module then infers the spatial attention map before generating the final haze-free image. Experimental results, including several quantitative metrics, demonstrate the efficiency and scalability of the suggested methodology.

\end{abstract}
%
%
\section{Introduction}
In inclement weather, a commonly observed atmospheric phenomenon is haze, which impedes visibility in natural scenes (Fig. 1). It stems from the absorption or scattering of light when it passes through an assemblage of tiny, suspended particles. As a result, the observed scene has artifacts like subdued brightness and anomalous contrast, hue, and saturation. Moreover, this problem has gained widespread attention for the implications it has on a model's performance in complex vision tasks like satellite imagery and autonomous driving \cite{zhu2021atmospheric,yin2021attentive,shin2021photo}. \par
The objective behind this work is to approximate a dehazed image given its hazy equivalent. In general, the haze-free image is estimated utilizing the atmospheric scattering model.  \cite{mccartney1976optics}. Mathematically:       
\begin{equation}
     I(x) = J(x)t(x) + A[1 - t(x)]
\end{equation}
\\
Here, for a pixel coordinate $x$, $I(x)$ is the observed image with haze, $J(x)$ is its haze-free equivalent, $A$ equals the global atmospheric light and $t(x)$ is the medium’s transmission matrix further put as:
\begin{equation}
    t(x)=e^{-\beta d(x)}
\end{equation}
\\
 $\beta$ being the coefficient of atmospheric scattering and $d(x)$ the depth of scene. Given the number of unknown parameters, the resultant problem is an ill-posed one. Traditionally, priors are used to approximate values for $A$ and $t(x)$ \cite{he2010single,fattal2014dehazing,zhu2015fast} or in certain experiments, $A$ is intuitively equated to a constant. Despite giving considerable results, the prior-based approaches are based on certain assumptions and estimations due to which, they at times do inadequate dehazing and introduce artifacts in the haze-free image such as unnatural contrast and color blockings. \par
Contrary to this, learning-based methods rely on a training process \cite{ren2018gated,dong2020multi,wang2021eaa} where the model, rather than relying on self-curated priors to estimate $A$ and $t(x)$, learns a mapping for the same, given a considerable number of hazy images and their dehazed equivalents. More recently, vision transformers (ViT) \cite{vaswani2017attention,dosovitskiy2020image} have transcended most of the traditional learning-based networks in high-level vision tasks. 
\begin{figure*}
    \centering
    \includegraphics[width=10cm]{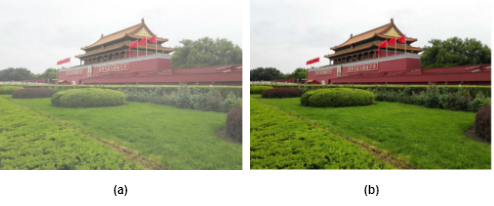}
    \caption{ (a) A hazy image. (b) Haze-free equivalent.}
\end{figure*}
\par
This work proposes an image-dehazing transformer in an attention network coupled with a CNN and a residual network that gives a haze-free image in the output. Input to the CNN is an image $I(x)$ affected by the haze. The network approximates a transmission matrix $t(x)$ as output whose ratio with the image affected by haze $(I(x)/ t(x))$ goes into the residual network as input, the output being the residue between the hazy input and the potential haze-free image. Pertinent to mention is that this network does not need the approximation of the global atmospheric light $A$ and so, is immune to the anomalies induced by it. Thereafter, an attention module employs transformer encoders that take into consideration both the global context and the scene depth of an image to infer the channel attributes of the residual image. Finally, the spatial attention is approximated, and the feature maps are concatenated to help estimate the final dehazed image.
\\
\section{Related Work}
A concise overview of the methodologies relevant to Image dehazing are discussed in this section. It is broadly classified into methods that rely on priors and learning-based methods. \par
The prior-based methods focus on calculating the transmission matrix and global atmospheric light by making certain presumptions about the scene and using manually customized priors. Tang \emph{et al.} \cite{tang2014investigating} try to investigate the best features that help remove the haze. They reach the conclusion that the dark-channel features are the most important ones when it comes to haze removal. Despite being trained on patches of synthetic images, the model does well on real-world data. He \emph{et al.} \cite{he2010single} estimate a dark-channel prior which, alongside an imaging model helps in estimating the depth of haze in a single image and its subsequent dehazing. Another key learning here is that one or a few color channels in a hazy image have pixels of low intensity. Also, the authors show how the patch size significantly alters the output. Fattal \emph{et al.} \cite{fattal2008single} came up with a novel haze removal model which, in addition to the transmission map, takes into account the surface shading too, based on an assumption that they are orthogonal. A similar hypothesis is used to determine the hue of haze. Despite using images from different sources, the model gives a mean absolute error of less than 7\% on dehazed output. Meng \emph{et al.} \cite{meng2013efficient} exploits the boundary limits of the transmission map along with the L1-norm-based contextual regularization to model an optimization problem for image dehazing. Also, this function is modeled on few assumptions about the scene making it more fault-tolerant. Zhao \emph{et al.} \cite{zhao2021single} divide the hazy image into smaller blocks followed by a patch-wise estimation of the transmission maps. Further optimization results in refined details and well-defined edges in the output image. Pertinent to mention is that since these methods are dependent on hand-crafted priors and scene constraints, they exhibit less fault tolerance and flexibility. \par
The widespread application of deep learning in image dehazing tasks has given noticeable results. In these learning-based methods, a model learns a set of parameters and an input-output mapping when trained on images and uses the learned parameters to dehaze the rest, usually by estimating a medium transmission map. In, authors Ren \emph{et al.} \cite{ren2016single} learn a mapping from a single hazy input image to its corresponding transmission matrix using a coarse-scale network. Following this, another neural network helps refine the haze-free output. Cai \emph{et al.} \cite{cai2016dehazenet} use a trainable CNN-based module to approximate a transmission map, whose layers are crafted in a way that they extract features for which priors were used. Finally, a Bilateral Rectified Linear Unit is used for output quality enhancement. The work proposed by Zhang \emph{et al.} in \cite{zhang2018densely} has an end-to-end image dehazing module where a DenseNet-inspired encoder-decoder helps approximate the transmission matrix while the method jointly infers the atmospheric light and the dehazing task with it. Although further employing a discriminator improves the quality of the haze-free images, the output generated has low contrast. Li \emph{et al.} \cite{li2018image} propose a residual learning-based network for haze removal. Here, the model first approximates a transmission map and then takes its ratio with the input image to generate the residual output. Also, the network functions without a need to estimate a value for the atmospheric light. Authors suggest it saves the model from the influence of invalid parameters. Despite this, the model is slow at processing and does inadequate dehazing of bright regions of the sky. In  \cite{gao2022novel}, Gao \emph{et al.} uses a CNN and a vision transformer for dehazing images. They claim that although CNNs perform better than prior-based methods, they lag behind vision transformers when it comes to detailed information retrieval. However, it does not take into consideration the global features of the image and performs weakly on dark scenes and varying depth. Li \cite{li2022two} \emph{et al.} propose a two-stage image dehazing module in which a Swin transformer along with a CNN is used in the first stage and the local features extracted in the second, while an attention mechanism is used between them. However, the computational cost to dehaze a real-world image is high. Also, the complexity of the network slows down the training too.  \par   
Even though learning-based methods outperform the prior-based ones, methods used to further improve the performance of learning-based networks add more depth to the network, resulting in adding more skip-connections, thereby increasing the network complexity and in certain cases, inducing noise too. As a result, by including a residual network, a scalable model is presented, which when used in conjunction with vision transformers, yields encouraging results. 

\section{Methodology}
The overall network architecture of the proposed methodology is shown in Figures 2 and 3. It dehazes the input image by passing it through two modules: the first one being a residual network and the other one a vision transformer-based attention module. Relevant to mention is that although the residual network alone can provide a dehazed image, the results are unsatisfactory since the output is not free of artifacts and color degradation. Also, it performs poorly on the quality metrics used in this work. Stacking it with an attention network enhances its dehazing performance, as will be evident from various quality metrics in the result section. The working of the two modules and the dataset description are discussed in this section.
\subsection{Dataset}\label{AA}
This work relies on the NYU2 depth dataset \cite{silberman2012indoor} for training the model and SOTS and HSTS datasets \cite{li2018benchmarking} for testing it. In practice, there is a shortage of pairs of hazy and dehazed images large enough to train a dehazing model efficiently. Pertinently, since the NYU2 dataset has clear haze-free images and their depth maps, we use these depth maps to get the transmission matrix and assign a constant value to the global atmospheric light to finally generate the required hazy dataset. We do this to 1000 such images from the NYU2 dataset, scaled down to 16 x 16 x 3 for faster training. Although the train and test images have been captured in indoor scenes with varying luminescence, the model performs equally well in outdoor setups. For testing the model, we take 400 and 20 images from \cite{li2018benchmarking} respectively. Here, the images in SOTS are captured in indoor setups while the ones in HSTS are captured outdoors. 
\begin{figure*}
    \centering
    \includegraphics[width= 1\textwidth]{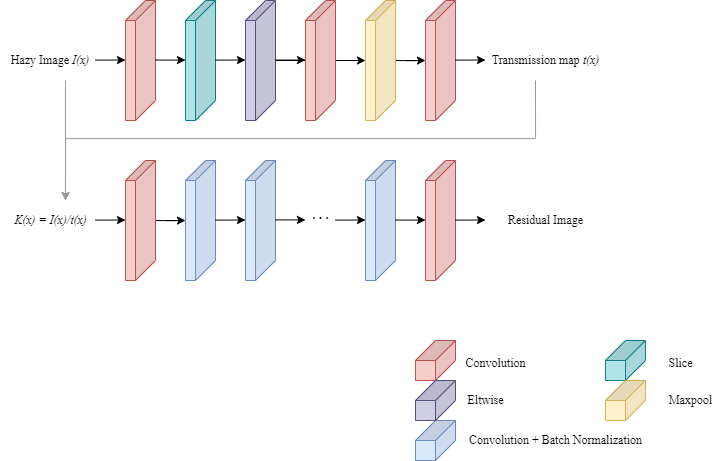}
    \caption{Network architecture of the residual module. A description of the dimensions of these layers is given in section 3.4. }
\end{figure*}
\subsection{The Residual Network}
Previously, work done on image dehazing had an increased focus on approximating both the transmission matrix $t(x)$ and the global atmospheric light $A$ to get $J(x)$. Erratic scene conditions make the estimation of these variables a tricky task. We solve equation (1) so that values for $A$ and $t(x)$ are learned by the network. This helps in avoiding the abnormalities the misestimation of A causes to the network architecture. Also, the benefit of residual learning is that the impact is superior to direct learning when residual learning is used to characterize the approximate identity mapping. Dividing equation (1) by $t(x)$ on both sides, we have: \par
\begin{equation}
    \frac{I(x)}{t(x)} =J(x)+\frac{A[1-t(x)]}{t(x)}
\end{equation}
Let K(x)=\(\frac{I(x)}{t(x)}\) and \(u(x)=\frac{A[1-t(x)]}{t(x)}\). 
\\
Equation (3) can be written as:
\begin{equation}
    K(x)=J(x)+u(x)
\end{equation}
In the proposed work, $I(x)$ passes through a CNN and the output is an approximation of the transmission matrix $t(x)$. Their ratio $K(x)$ is the input to the residual network with the aim to generate a residual image $(R'(K(x)))$. The haze-free image $J(x)$ can be given as:
\begin{equation}
    J(x)=K(x)-R'(K(x))
\end{equation}
Here, \(R'(K(x))\approx u(x)\).
\\
The loss function is described as:
\begin{equation}
    l(\theta )= \frac{1}{2n}\sum_{i=1}^{n}\left \| R'(K(x_{i}))-(K(x_{i})-J(x_{i})) \right \|_{F}^{2}
\end{equation}

This function trains $\theta$ to give the average squared error between the residual image estimated by the network $R'(K(x))$ and the expected one $(K(x)-J(x))$. The residual image simultaneously goes through two convolution layers with kernel sizes $3 \times 3$ and $5 \times 5$ giving 3 and 9 output channels respectively. These channels are concatenated with the residual image and the output generated goes into the attention module.
\begin{figure*}
    \centering
    \includegraphics[width= 1\textwidth]{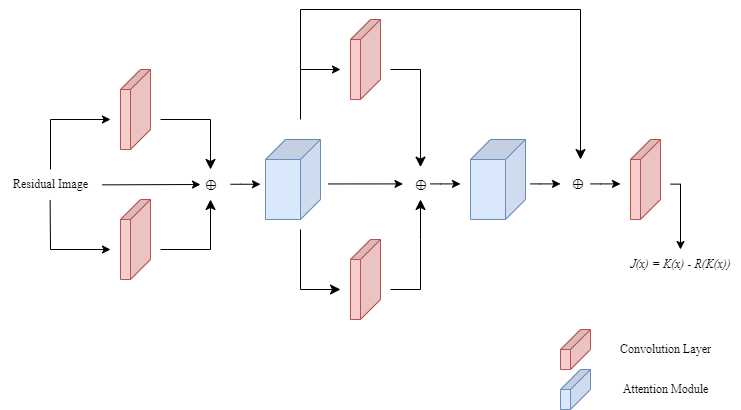}
    \caption{Graphical overview of the proposed architecture. $\oplus$  denotes channel concatenation.}
\end{figure*}
\subsection{The Attention Module}
The attention network proposed in this work takes inspiration from CBAM \cite{woo2018cbam} which successively infers channel and spatial attention. However, in order to generate these attention maps, CBAM relies on operations such as max-pooling and average-pooling. We approach the task of inferring spatial attention with a vision transformer to demonstrate its effectiveness. The aim of using an attention module is to enhance and further dehaze the output $(R'(K(x))$ from the residual network in a way that takes into consideration both the global context and scene depth in an image. \par
The transformer here takes as input a sequence of one-dimensional tokens. The concatenated output in the previous step is divided into two-dimensional patches of resolution m x m which are then flattened into one-dimensional tokens. Mathematically, an image with $C$ channels and dimensions $H\times W$ given by \(x \in R^{C\times H\times W}\), is flattened into patches \(x_{m} \in R^{N\times (m^{2}.C)}\)  where $N$ gives the total number of such patches. Since the latent vector size used here does not vary across layers, all these tokens are linearly projected to a constant dimension. Furthermore, to preserve the position information of these patches, learnable position embeddings \(E_{t}\) are used. The elaborated architecture of the transformer employed is given in Fig. 4. The final vector input to the transformer is: 
\begin{equation}
    T_{i} = x_{m}+E_{t}
\end{equation}
\par
\begin{figure*}
    \centering
    \includegraphics[width= 1\textwidth]{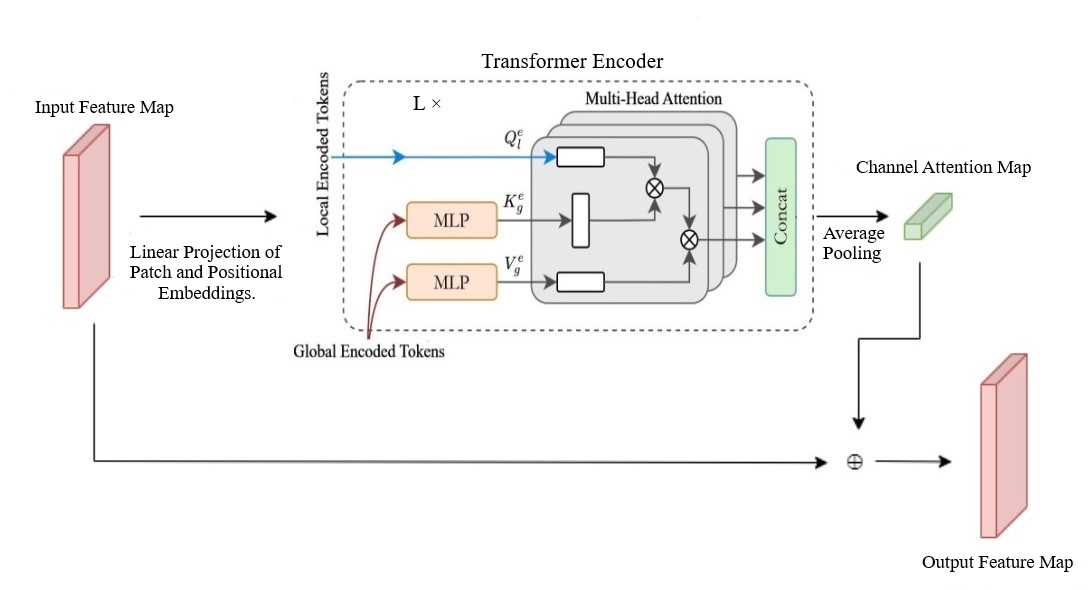}
    \caption{The transformer-based channel attention module.}
\end{figure*}
The transformer encoder comprises multi-head attention (MHA) and multilayer perceptron (MLP) blocks embedded between alternating layers of normalization (LN). For an $nth$ transformer:
\begin{equation}
    T'_{n-1} = MHA(LN(T_{n-1}))+T_{n-1}; n=1 \dots N,
\end{equation}
\begin{equation}
    T_{n} = MLP(LN(T'_{n-1}))+T'_{n-1},     n=1 \dots N.
\end{equation}
\par
Our approach utilizes global context self-attention modules in conjunction with conventional local self-attention mechanisms, enabling the effective and efficient modeling of spatial interactions across both long and short ranges. While doing so, it also learns to effectively dehaze at varying depth levels as well. As shown in Figure 4, the local token embeddings are generated from the patch projections of the input feature map. For generating the global tokens, we take inspiration from \cite{hatamizadeh2023global}. The network then exploits the local queries for generic encoder operations with global keys and values. Finally, average pooling over the output tokens gives the output as a channel attention map. Further, max-pooling and average-pooling help generate the spatial attention map in a way similar to CBAM. The feature maps generated by channel and spatial attention modules are concatenated to obtain $(R(K(x))$ before a final convolution operation over their feature maps. All convolution operations are performed as shown in Figure 3. The output here $(R(K(x))$ is subtracted from $K(x)$ to get the final haze-free output $(J(x))$. Experimental results demonstrate the distinction of this method. 
\subsection{Implementation Details}
Because of the dearth of a considerable number of hazy and haze-free image pairs, the NYU2 depth dataset was used to estimate hazy images from depth maps and haze-free images. A constant global atmospheric light value along with a transmission map approximated from scene depth helped with this estimation. Input to the model is a hazy image $I(x)$ of size $16 \times 16 \times 3$. The feature map obtained after passing through the first convolution layer is of size $14 \times 14 \times 16$. Further, the slice and eltwise layers convert it to feature maps of size $14 \times 14 \times 4 \times 4$ and $14 \times 14 \times 4$ respectively. A $7 \times 7$ max-pooling layer and ReLu activation is further applied to produce the transmission map $t(x)$. In the lower branch, ReLu is applied after every layer. 13 Convolution + Batch Normalization layers are used here, with $3 \times 3$ filters applied throughout. Output from the residual network is passed through convolution layers of kernel size $3 \times 3$ and $5 \times 5$ simultaneously, the feature maps concatenated with the residual image before being passed through the attention module, the detailed architecture of which can be found in Figure 4. Output from the first transformer is further concatenated with layers that use $3 \times 3$ (same for the final layer) and $5 \times 5$ kernels. The final enhanced residual output is then subtracted from $K(x)$ to get the haze-free output.

\section{Results}
The proposed architecture is trained on 1000 images from the NYU2 dataset on a single Nvidia RTX 3090 Ti GPU for 150 epochs and a batch size of 16. A number of data augmentation techniques like random crop, horizontal flipping, and random rotation were applied to avoid overfitting.  For testing the model, we take 400 and 20 images from SOTS and HSTS respectively. The outcomes are contrasted with various state-of-the-art image dehazing techniques like Fattal’s \cite{fattal2008single}, DehazeNet \cite{cai2016dehazenet} and Gao \emph{et al.} \cite{gao2022novel}. Along with the visual results, three metrics are used to evaluate the restored haze-free images' quality and compare the relative performance of different methods. These include PSNR \cite{hore2010image}, SSIM \cite{wang2004image}, and FSIM \cite{zhang2011fsim}. The better performance of a restoration method is judged by higher values of these spatial metrics. Mathematically, PSNR, SSIM, and FSIM are defined as follows:
\begin{equation}
    PSNR(x,y)= 10\log_{10}(\frac{MAX^{2}}{MSE(x,y)})
\end{equation}
\par
Here $MSE$ is the mean squared error between images $x$ and $y$ and $MAX$ equals the maximum possible value of the pixels in it.
\begin{equation}
    SSIM(x,y)=l(x,y)c(x,y)s(x,y)
\end{equation}
where, respectively, $l(x, y)$, $c(x, y)$, and $s(x, y)$ quantify the distance between the two images' mean luminance, mean contrast, and correlation coefficient.
\begin{equation}
    FSIM=\frac{\sum a\in\omega S_{L}(a).PC_{m}(a)}{\sum a\in\omega PC_{m}(a)}
\end{equation}
\par
The maximum phase congruency for a location $a$ is represented by the component $PC_{m}(a)$, and $\omega$ represents the spatial domain of the entire image. Tables 1 and 2 give the quantitative evaluation using the mean of PSNR, SSIM and FSIM over the SOTS and HSTS datasets respectively.
\begin{table}[ht!]
\centering
\caption{Mean PSNR, SSIM, and FSIM of the given methods over the SOTS dataset.}
\begin{tabular}{ c c c c c }
\hline
 Metrics & Fattal's & DehazeNet & Gao \emph{et al.} & Proposed \\ 
 \hline
 MPSNR & 16.74 & 15.49 & 20.86 & 22.93 \\
 MSSIM & 0.705 & 0.739 & 0.874 & 0.903 \\
MFSIM & 0.946 & 0.884 & 0.949 & 0.933 \\

 \hline
 \\
\end{tabular}
\end{table}

\begin{table}[ht!]
\centering
\caption{Mean PSNR, SSIM, and FSIM of the given methods over the HSTS dataset.}
\begin{tabular}{ c c c c c }
\hline
 Metrics & Fattal's & DehazeNet & Gao \emph{et al.} & Proposed \\ 
\hline
MPSNR &	17.74 &	22.39 &	23.26 &	26.83 \\
MSSIM &	0.803 &	0.817 &	0.917 &	0.924 \\
MFSIM &	0.901 &	0.952 &	0.921 &	0.978 \\

\hline
 \\
\end{tabular}
\end{table}
\subsection{Ablation Study}
To corroborate the efficacy of the proposed network, an ablation study was designed in which the network was trained with and without the transformer-based attention module on the HSTS dataset. The test set was used to calculate the mean SSIM and PSNR of the architectures thus formed. The proposed module reported SSIM and PSNR that were 6.27\% and 11.43\% higher than the model which used just the residual module. This study thus demonstrates the importance of using an attention network alongside a residual module for effectual image dehazing.
\begin{figure*}
    \centering
    \includegraphics[width= 1\textwidth]{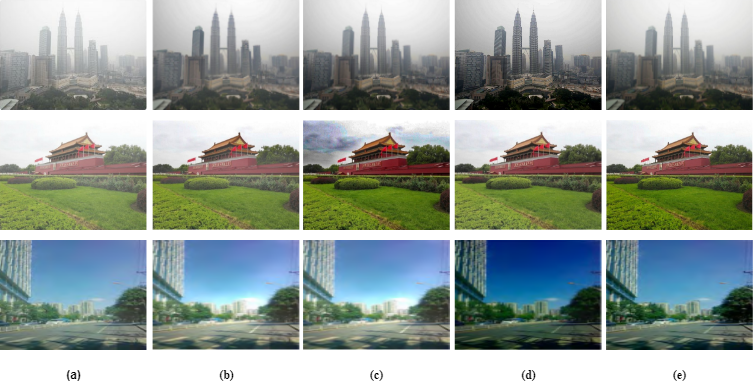}
    \caption{Dehazing results on the HSTS Dataset. (a) Hazy Image; (b) Fattal’s; (c) DehazeNet; (d) Gao \emph{et al.}; (e) Proposed Network.}
\end{figure*}

\section{Conclusion and Future Scope}
This work proposed a residual learning transformer in an attention module-based image dehazing network. Usually, work done on obtaining haze-free images relies on approximating the transmission matrix and atmospheric light out of the atmospheric scattering model equation. Estimating these variables in an ill-posed problem makes it prone to misestimation. In this paper, the residual module learns these variables. The channel attention network and the pooled spatial maps from the transformer further enhance the performance of the residual backbone. Experiments done on the NYU2, SOTS, and HSTS datasets demonstrate the robustness of the proposed methodology as can be validated by both visual results and quantitative evaluation given using PSNR, SSIM, and FSIM. Not only does the method estimate output with a high degree of similarity with the ground truth, but it also does so without inducing any artifacts. In the future, we intend to generate a large dataset captured in both indoor and outdoor setups with more diverse lighting conditions to train a network robust enough to do real-time image dehazing.  

\bibliographystyle{ieeetr}
\bibliography{references.bib}
\end{document}